\documentclass[conference]{IEEEtran}
\IEEEoverridecommandlockouts
\usepackage{cite}
\usepackage{amsmath,amssymb,amsfonts}
\usepackage{algorithmic}
\usepackage{url}
\usepackage{multirow}
\usepackage{graphicx}
\usepackage{textcomp}
\usepackage{xcolor}

\usepackage{cleveref}
\newcommand\BibTeX{B\textsc{ib}\TeX}
\crefformat{section}{\S#2#1#3}
\crefformat{subsection}{\S#2#1#3}
\crefformat{subsubsection}{\S#2#1#3}

\def\BibTeX{{\rm B\kern-.05em{\sc i\kern-.025em b}\kern-.08em
    T\kern-.1667em\lower.7ex\hbox{E}\kern-.125emX}}
\begin{document}

\title{AST-Transformer: Encoding Abstract Syntax Trees Efficiently for Code Summarization}

\author{
\IEEEauthorblockN{Ze Tang\textsuperscript{1}, Chuanyi Li\textsuperscript{1}, Jidong Ge\textsuperscript{1}, Xiaoyu Shen\textsuperscript{2}, Zheling Zhu\textsuperscript{1}, Bin Luo\textsuperscript{1}} 

\IEEEauthorblockA{
		\textsuperscript{1}The State Key Laboratory for Novel Software Technology, Nanjing University\\
		\textsuperscript{2}Saarland Informatics Campus\\
}

}

\maketitle

\begin{abstract}
Code summarization aims to generate brief natural language descriptions for source code.
As source code is highly structured and follows strict programming language grammars, its Abstract Syntax Tree (AST) is often leveraged to inform the encoder about the structural information.
However, ASTs are usually much longer than the source code. Current approaches ignore the size limit and simply feed the whole linearized AST into the encoder.
To address this problem, we propose AST-Transformer to efficiently encode tree-structured ASTs. 
Experiments show that AST-Transformer outperforms the state-of-arts by a substantial margin while being able to reduce $90\sim95\%$ of the computational complexity in the encoding process.
\end{abstract}


\section{Introduction}
The summary of source code is a brief natural language description explaining the purpose of the code~\cite{code_sum_java_methods_2016}.
The code to be summarized can be with different units. In this work, we focus on summarizing the subroutines or defined methods in a program.

Current state-of-the-arts all follow the encoder-decoder architecture~\cite{shen2017estimation,shen2019improving,Vaswani2017} and can be trained end-to-end with code-summary pairs. Since the source code is highly structured and follows rigid programming language grammars, a common practice is to also leverage the Abstract Syntax Tree (AST) to help the encoder digest the structured information. The AST is usually linearized by different algorightms like pre-order traversal~\cite{codePredict2020}, structure-based traversal (SBT)~\cite{DeepCom2018} and path decomposition~\cite{code2seq2019}, then fed into the encoder. Several works also proposed architectures specific for tree encoding like tree-LSTM~\cite{Eriguchi2016,code_sum_tree_lstm_reinforce_2018}.

However, the linearized ASTs, as containing additional structured information, are much longer than their corresponding source code sequence. Some linearization algorithms can further increase the length. For example, linearizing with SBT usually doubles the size of original AST. This makes the model extremely difficult to accurately detect useful dependency relations from the overlong input sequence
. Moreover, it brings significant computational overhead, especially for those state-of-the-art Transformer-based models where the number of self-attention operations grows quadratically with the sequence length. Encoding ASTs with tree-based models like tree-LSTM will incur extra complexity as they need to traverse the whole tree to obtain the state of each node.

In this work, we argue that
it is unnecessary to model the dependency between every single node pair. Our intuition is that the state of a node in the AST is affected most by its (1) ancestor-descendent nodes, which represent the hierarchical relationship within one operation, and (2) sibling nodes, which represent the temporal relationship across different operations. Based on this intuition, we propose AST-Transformer, a simple variant of the Transformer model to efficiently handle the tree-structured AST.

\section{Approach}
This section details our proposed AST-Transformer, i.e., a simple yet effective Transformer variant to deal with the tree-structured AST.
The overall architecture of AST-Transformer has two main parts, i.e., AST Encoder and Decoder respectively. The particularity of the proposed AST-Transformer lies in the three special components in the Encoder, namely, AST Linearization, Relation Matrices, and Tree-Structure Attention. First, in subsection \cref{sec: linearization}, three different linearization methods for transforming input AST into a sequence are introduced. Then, two matrices for encoding the ancestor-descendent and sibling relationships in the tree are defined in subsection \cref{sec: relation_matrices}, as well as concrete methods for constructing the matrices. Eventually, the proposed self-attention mechanism based on relation matrix for generating code summaries is illustrated in subsection \cref{sec: tree-attention}. 

\begin{table*}[ht]
\caption{Comparison of AST-Transformer with the baseline methods}
\begin{center}
\resizebox{\linewidth}{!}{%
\begin{tabular}{l|c|ccc|ccc}
\hline
\multirow{2}{*}{Methods} & \multirow{2}{*}{Input} & \multicolumn{3}{c|}{Java dataset~\cite{code_sum_with_api_2018}}                        & \multicolumn{3}{c}{Python dataset\cite{code_sum_tree_lstm_reinforce_2018}}                      \\ \cline{3-8}                                 &                        & BLEU(\%)          & METEOR(\%)         & ROUGE-L(\%)        & BLEU(\%)           & METEOR(\%)         & ROUGE-L(\%)        \\ \hline
CODE-NN\cite{code_sum_cnn_2016}                       &                        & 27.6           & 12.61          & 41.10          & 17.36          & 9.29          & 37.81          \\
Dual Model\cite{dual_task_2019}                       & Code                   & 42.39          & 25.77          & 53.61          & 21.80          & 11.14          & 39.45          \\
Transformer(Code)\cite{code_sum_trans_2020}           &                        & 44.58          & 26.43          & 54.76          & 32.52          & 19.77          & 46.73          \\ \hline
DeepCom\cite{DeepCom2018}                             &                        & 39.75          & 23.06          & 52.67          & 20.78          & 09.98          & 37.35          \\
API+CODE\cite{code_sum_with_api_2018}                 & SBT                    & 41.31          & 23.73          & 52.25          & 15.36          & 08.57          & 33.65          \\
Transformer(SBT)*                                     &                        & 39.76          & 24.46          & 53.64          & 21.02          & 10.48          & 38.42          \\ \hline
Code2Seq*\cite{code2seq2019}                          & PD                     & 24.42          & 15.35          & 33.95          & 17.54          & 08.49          & 20.93          \\ \hline
Tree2Seq\cite{Eriguchi2016}                           & \multirow{2}{*}{AST}   & 37.88          & 22.55          & 51.50          & 20.07          & 08.96          & 35.64          \\
RL+Hybrid2Seq\cite{code_sum_tree_lstm_reinforce_2018} &                        & 38.22          & 22.75          & 51.91          & 19.28          & 09.75          & 39.34          \\ \hline
AST-Transformer                                       & POT + R                & \textbf{46.64} & \textbf{28.49} & \textbf{55.21} & \textbf{33.82} & \textbf{21.96} & \textbf{47.14} \\ \hline
\end{tabular}
}
\label{table:overall_results}
\end{center}
\end{table*}

\subsection{AST Linearization}
\label{sec: linearization}
In order to make the tree-shaped AST suitable as the input of the neural network model, it first needs to be converted into a sequence with a linearization method. Technically, the proposed AST-Transformer is orthogonal to the linearization and can be build upon any concrete approach. In this paper, the three most representative methods are selected for conducting experiments to test which one can achieve the best effect in combination with the self-attention based on the relation matrix, and they are: Pre-order Traversal (POT), Structure-based Traversal (SBT)~\cite{DeepCom2018} and Path Decomposition (PD)~\cite{code2seq2019}.

We find that the performances of using SBT and PD have no big differences with using POT in AST-Transformer through experiments. However, generating POT saves almost $90\sim95\%$ time costs compared with generating SBT or PD for the entire dataset. And fortunately, using the simplest POT has been able to achieve the state-of-the-art performance. 

\subsection{Relationship Matrices}
\label{sec: relation_matrices}
We define two kinds of relationships between nodes in the tree that we care about: the ancestor-descendant relationship and the sibling relationship. The former represents the hierarchical information within one operation, and the latter represents the temporal information across different operations.
Specifically, two nodes have the ancestor-descendant relationship if there exists a directed path from root node that can traverse through them.
Two nodes have the sibling relationship if they share the same parent node.

We use two matrices, i.e., $A_{N \times N}$ and $S_{N \times N}$, to represent the ancestor-descendent and sibling relationships respectively. $N$ is the total number of nodes. We denote the $i$th node in the linearized AST as $n_i$. $A_{i,j}$ is the distance of the shortest path between $n_i$ and $n_j$ in the AST. $S_{i,j}$ is horizontal sibling distance between $n_i$ and $n_j$ in the AST if they satisfy the sibling relationship. If one relationship is not satisfied, its value in the matrix will be infinity. 

By taking the advantages of these two matrices, the model can find related nodes in parallel and efficiently by scanning the matrices instead of traversing the initial tree. 

\subsection{Tree-Structure Attention}
\label{sec: tree-attention}
For incorporating with the relationship matrices in self-attention, we combine the practices of Shaw \textit{et al.} \cite{relative_position_rep_2018} and He \textit{et al.} \cite{deberta2020}. 
Similar to the vanilla Transformer, we use multi-head attention to jointly attend to information from different relationship matrices.
Then the outputs of the self-attention with the ancestor-descendant and the sibling relationship matrices are concatenated and once again projected, resulting in the final values.

\section{EXPERIMENTS}
The overall result of AST-Transformer and the baselines are proposed in Table~\ref{table:overall_results}.
Results show that AST-Transformer obviously outperforms all the baselines in all three metrices. 
AST-Transformer outperforms the nearest baseline using code token sequence as input by 2.06, 1.3 BLEU, 2.04, 2.19 METEOR and 0.45, 0.41 ROUGE-L in the Java and Python datasets respectively.
And the improvement is more obvious compared with baselines using AST or linearized AST as input.
We think there are two main reasons for the improvement of AST-Transformer.
Firstly, top two approaches (AST-Transformer and Transformer(CODE)) both use the Transformer architecture.
As the length of code or AST is much longer than the natural language,  self-attention mechanism is conductive to help the model catch long distance meaningful word pair or node pair, and then learn some characters related to the code function.
Secondly, though we say that AST contains more information than code token sequence, as AST not only has the semantic information (which is stored in leaf nodes), but also has the structural information (which is stored in non-leaf nodes), the performances of most approaches based on AST are inferior to Transformer(Code), a model just using code token sequence as input.
It may be illustrated by that there are many very general structures, such as MethodDeclare $\rightarrow$ MethodBody, in AST.
These structures basically occur in every code, and they are noisy information for models, just like the pause words in nature language.
This is exactly why Transformer(SBT) has hardly improved compared to DeepCom, as SBT has around 4 times more nodes than AST.
In AST-Transformer, we only allow the node exchanges information with other nodes that are no more than $K$ away from it.
This can effectively ensure the specificity of each node without being assimilated by the overall structure.

\section{Conclusion}
In this paper, we have presented a new Transformer-based model that can encode AST effectively.
By using two relationship matrices, AST-Transformer can encode AST without suffering from the computational complexity.
Comprehensive experiments show that AST-Transformer outperforms other competitive baselines and achieves the state-of-art performance on several automatic metrics.
\section{ACKNOWLEDGMENTS}
 \emph{This work is supported by National Natural Science Foundation of China (61802167,61972197,61802095) ,Natural Science Foundation of Jiangsu Province (No.BK20201250),and the Cooperation Fund of Huawei-NJU Creative Lab for the Next Programming.}
\bibliographystyle{IEEEtran}
\bibliography{main}

\begin{thebibliography}{10}
\providecommand{\url}[1]{#1}
\csname url@samestyle\endcsname
\providecommand{\newblock}{\relax}
\providecommand{\bibinfo}[2]{#2}
\providecommand{\BIBentrySTDinterwordspacing}{\spaceskip=0pt\relax}
\providecommand{\BIBentryALTinterwordstretchfactor}{4}
\providecommand{\BIBentryALTinterwordspacing}{\spaceskip=\fontdimen2\font plus
\BIBentryALTinterwordstretchfactor\fontdimen3\font minus
  \fontdimen4\font\relax}
\providecommand{\BIBforeignlanguage}[2]{{%
\expandafter\ifx\csname l@#1\endcsname\relax
\typeout{** WARNING: IEEEtran.bst: No hyphenation pattern has been}%
\typeout{** loaded for the language `#1'. Using the pattern for}%
\typeout{** the default language instead.}%
\else
\language=\csname l@#1\endcsname
\fi
#2}}
\providecommand{\BIBdecl}{\relax}
\BIBdecl

\bibitem{code_sum_java_methods_2016}
\BIBentryALTinterwordspacing
P.~W. McBurney and C.~McMillan, ``Automatic source code summarization of
  context for java methods,'' \emph{{IEEE} Trans. Software Eng.}, vol.~42,
  no.~2, pp. 103--119, 2016. [Online]. Available:
  \url{https://doi.org/10.1109/TSE.2015.2465386}
\BIBentrySTDinterwordspacing

\bibitem{shen2017estimation}
X.~Shen, Y.~Oualil, C.~Greenberg, M.~Singh, and D.~Klakow, ``Estimation of gap
  between current language models and human performance,'' \emph{Proc.
  Interspeech 2017}, pp. 553--557, 2017.

\bibitem{shen2019improving}
X.~Shen, Y.~Zhao, H.~Su, and D.~Klakow, ``Improving latent alignment in text
  summarization by generalizing the pointer generator,'' in \emph{Proceedings
  of the 2019 Conference on Empirical Methods in Natural Language Processing
  and the 9th International Joint Conference on Natural Language Processing
  (EMNLP-IJCNLP)}, 2019, pp. 3762--3773.

\bibitem{Vaswani2017}
\BIBentryALTinterwordspacing
A.~Vaswani, N.~Shazeer, N.~Parmar, J.~Uszkoreit, L.~Jones, A.~N. Gomez,
  L.~Kaiser, and I.~Polosukhin, ``Attention is all you need,'' in
  \emph{Advances in Neural Information Processing Systems 30: Annual Conference
  on Neural Information Processing Systems 2017, 4-9 December 2017, Long Beach,
  CA, {USA}}, I.~Guyon, U.~von Luxburg, S.~Bengio, H.~M. Wallach, R.~Fergus,
  S.~V.~N. Vishwanathan, and R.~Garnett, Eds., 2017, pp. 5998--6008. [Online].
  Available: \url{http://papers.nips.cc/paper/7181-attention-is-all-you-need}
\BIBentrySTDinterwordspacing

\bibitem{codePredict2020}
\BIBentryALTinterwordspacing
S.~Kim, J.~Zhao, Y.~Tian, and S.~Chandra, ``Code prediction by feeding trees to
  transformers,'' \emph{CoRR}, vol. abs/2003.13848, 2020. [Online]. Available:
  \url{https://arxiv.org/abs/2003.13848}
\BIBentrySTDinterwordspacing

\bibitem{DeepCom2018}
\BIBentryALTinterwordspacing
X.~Hu, G.~Li, X.~Xia, D.~Lo, and Z.~Jin, ``Deep code comment generation,'' in
  \emph{Proceedings of the 26th Conference on Program Comprehension, {ICPC}
  2018, Gothenburg, Sweden, May 27-28, 2018}, F.~Khomh, C.~K. Roy, and
  J.~Siegmund, Eds.\hskip 1em plus 0.5em minus 0.4em\relax {ACM}, 2018, pp.
  200--210. [Online]. Available: \url{https://doi.org/10.1145/3196321.3196334}
\BIBentrySTDinterwordspacing

\bibitem{code2seq2019}
\BIBentryALTinterwordspacing
U.~Alon, S.~Brody, O.~Levy, and E.~Yahav, ``code2seq: Generating sequences from
  structured representations of code,'' in \emph{7th International Conference
  on Learning Representations, {ICLR} 2019, New Orleans, LA, USA, May 6-9,
  2019}.\hskip 1em plus 0.5em minus 0.4em\relax OpenReview.net, 2019. [Online].
  Available: \url{https://openreview.net/forum?id=H1gKYo09tX}
\BIBentrySTDinterwordspacing

\bibitem{Eriguchi2016}
\BIBentryALTinterwordspacing
A.~Eriguchi, K.~Hashimoto, and Y.~Tsuruoka, ``Tree-to-sequence attentional
  neural machine translation,'' in \emph{Proceedings of the 54th Annual Meeting
  of the Association for Computational Linguistics, {ACL} 2016, August 7-12,
  2016, Berlin, Germany, Volume 1: Long Papers}.\hskip 1em plus 0.5em minus
  0.4em\relax The Association for Computer Linguistics, 2016. [Online].
  Available: \url{https://doi.org/10.18653/v1/p16-1078}
\BIBentrySTDinterwordspacing

\bibitem{code_sum_tree_lstm_reinforce_2018}
\BIBentryALTinterwordspacing
Y.~Wan, Z.~Zhao, M.~Yang, G.~Xu, H.~Ying, J.~Wu, and P.~S. Yu, ``Improving
  automatic source code summarization via deep reinforcement learning,'' in
  \emph{Proceedings of the 33rd {ACM/IEEE} International Conference on
  Automated Software Engineering, {ASE} 2018, Montpellier, France, September
  3-7, 2018}, M.~Huchard, C.~K{\"{a}}stner, and G.~Fraser, Eds.\hskip 1em plus
  0.5em minus 0.4em\relax {ACM}, 2018, pp. 397--407. [Online]. Available:
  \url{https://doi.org/10.1145/3238147.3238206}
\BIBentrySTDinterwordspacing

\bibitem{code_sum_with_api_2018}
\BIBentryALTinterwordspacing
X.~Hu, G.~Li, X.~Xia, D.~Lo, S.~Lu, and Z.~Jin, ``Summarizing source code with
  transferred {API} knowledge,'' in \emph{Proceedings of the Twenty-Seventh
  International Joint Conference on Artificial Intelligence, {IJCAI} 2018, July
  13-19, 2018, Stockholm, Sweden}, J.~Lang, Ed.\hskip 1em plus 0.5em minus
  0.4em\relax ijcai.org, 2018, pp. 2269--2275. [Online]. Available:
  \url{https://doi.org/10.24963/ijcai.2018/314}
\BIBentrySTDinterwordspacing

\bibitem{code_sum_cnn_2016}
\BIBentryALTinterwordspacing
M.~Allamanis, H.~Peng, and C.~Sutton, ``A convolutional attention network for
  extreme summarization of source code,'' \emph{CoRR}, vol. abs/1602.03001,
  2016. [Online]. Available: \url{http://arxiv.org/abs/1602.03001}
\BIBentrySTDinterwordspacing

\bibitem{dual_task_2019}
B.~Wei, G.~Li, X.~Xia, Z.~Fu, and Z.~Jin, ``Code generation as a dual task of
  code summarization,'' in \emph{Advances in Neural Information Processing
  Systems 32: Annual Conference on Neural Information Processing Systems 2019,
  NeurIPS 2019, December 8-14, 2019, Vancouver, BC, Canada}, H.~M. Wallach,
  H.~Larochelle, A.~Beygelzimer, F.~d'Alch{\'{e}}{-}Buc, E.~B. Fox, and
  R.~Garnett, Eds., 2019, pp. 6559--6569.

\bibitem{code_sum_trans_2020}
\BIBentryALTinterwordspacing
W.~U. Ahmad, S.~Chakraborty, B.~Ray, and K.~Chang, ``A transformer-based
  approach for source code summarization,'' in \emph{Proceedings of the 58th
  Annual Meeting of the Association for Computational Linguistics, {ACL} 2020,
  Online, July 5-10, 2020}, D.~Jurafsky, J.~Chai, N.~Schluter, and J.~R.
  Tetreault, Eds.\hskip 1em plus 0.5em minus 0.4em\relax Association for
  Computational Linguistics, 2020, pp. 4998--5007. [Online]. Available:
  \url{https://doi.org/10.18653/v1/2020.acl-main.449}
\BIBentrySTDinterwordspacing

\bibitem{relative_position_rep_2018}
\BIBentryALTinterwordspacing
P.~Shaw, J.~Uszkoreit, and A.~Vaswani, ``Self-attention with relative position
  representations,'' in \emph{Proceedings of the 2018 Conference of the North
  American Chapter of the Association for Computational Linguistics: Human
  Language Technologies, NAACL-HLT, New Orleans, Louisiana, USA, June 1-6,
  2018, Volume 2 (Short Papers)}, M.~A. Walker, H.~Ji, and A.~Stent, Eds.\hskip
  1em plus 0.5em minus 0.4em\relax Association for Computational Linguistics,
  2018, pp. 464--468. [Online]. Available:
  \url{https://doi.org/10.18653/v1/n18-2074}
\BIBentrySTDinterwordspacing

\bibitem{deberta2020}
\BIBentryALTinterwordspacing
P.~He, X.~Liu, J.~Gao, and W.~Chen, ``Deberta: Decoding-enhanced {BERT} with
  disentangled attention,'' \emph{CoRR}, vol. abs/2006.03654, 2020. [Online].
  Available: \url{https://arxiv.org/abs/2006.03654}
\BIBentrySTDinterwordspacing

\end{thebibliography}

\end{document}